\documentclass[runningheads,a4paper]{llncs}

\usepackage{amssymb}
\usepackage{graphicx}

\usepackage{algorithm}
\usepackage{algorithmic}
\usepackage{amsmath}

\usepackage{url}
\urldef{\mailsa}\path|arash.shahriari@{anu.edu.au, csiro.au}|    
\newcommand{\keywords}[1]{\par\addvspace\baselineskip
\noindent\keywordname\enspace\ignorespaces#1}

\begin{document}

\mainmatter  

\title{Multipartite Ranking-Selection of Low-Dimensional Instances by Supervised Projection to High-Dimensional Space}

\titlerunning{Multipartite Ranking-Selection of Low-Dimensional Instances}

%
%
\author{Arash Shahriari}
\authorrunning{}

\institute{Australian National University (ANU)\\
\& Commonwealth Scientific and Industrial Research Organisation (CSIRO)%
\\
\mailsa\\
\url{http://www.anu.edu.au}, \url{http://csiro.au}}

%
%

\toctitle{}
\tocauthor{}
\maketitle

\begin{abstract}

 Pruning of redundant or irrelevant instances of data is a key to every successful solution for pattern recognition. In this paper, we present a novel ranking-selection framework for low-length but highly correlated instances. Instead of working in the low-dimensional instance space, we learn a supervised projection to high-dimensional space spanned by the number of classes in the dataset under study. Imposing higher distinctions via exposing the notion of labels to the instances, lets to deploy one versus all ranking for each individual classes and selecting quality instances via adaptive thresholding of the overall scores. To prove the efficiency of our paradigm, we employ it for the purpose of texture understanding which is a hard recognition challenge due to high similarity of texture pixels and low dimensionality of their color features. Our experiments show considerable improvements in recognition performance over other local descriptors on several publicly available datasets.

\end{abstract}

\section{Introduction}

The literature on instance ranking and selection is vast. Sampling is a conventional method relying on random selection to form a subset of data. Another type of methods are based on selecting a set of relevant data in form of critical points, boundary points, prototypes and so on. They highly try to separate groups of data, or best represent each group~\cite{liu2013instance}. In contrast to the classification, instance ranking typically produces a ranking of instances by assigning a score to each instance and then sorting them by scores~\cite{furnkranz2011preference}. 

Instance ranking consists of two main proposals in preference learning which are bipartite and multipartite rankings~\cite{rajaram2005generalization}\cite{furnkranz2009binary}. They generally count the number of ranking errors such that bipartite amounts to the area under ROC curve~\cite{bradley1997use} or equivalently Wilcoxon statistic~\cite{wilcoxon1945individual} and multipartite generalizes to concordance index in statistics which is used to evaluate the discriminatory power and the predictive accuracy of nonlinear statistical models~\cite{gonen2005concordance}.

In this work, we propose a multipartite instance ranking-selection framework for low-dimensional instances which usually seem dense and highly similar because of their short lengths. This novel paradigm employs supervised learning to project instances to a high-dimensional space spanned by the number of classes in the dataset at hand. As a result, the notion of labels transfers to projected instances in the new space which imposes higher distinction among several classes. 

\begin{figure}[tb]
\begin{center}
\centerline{\includegraphics[width=12.5cm]{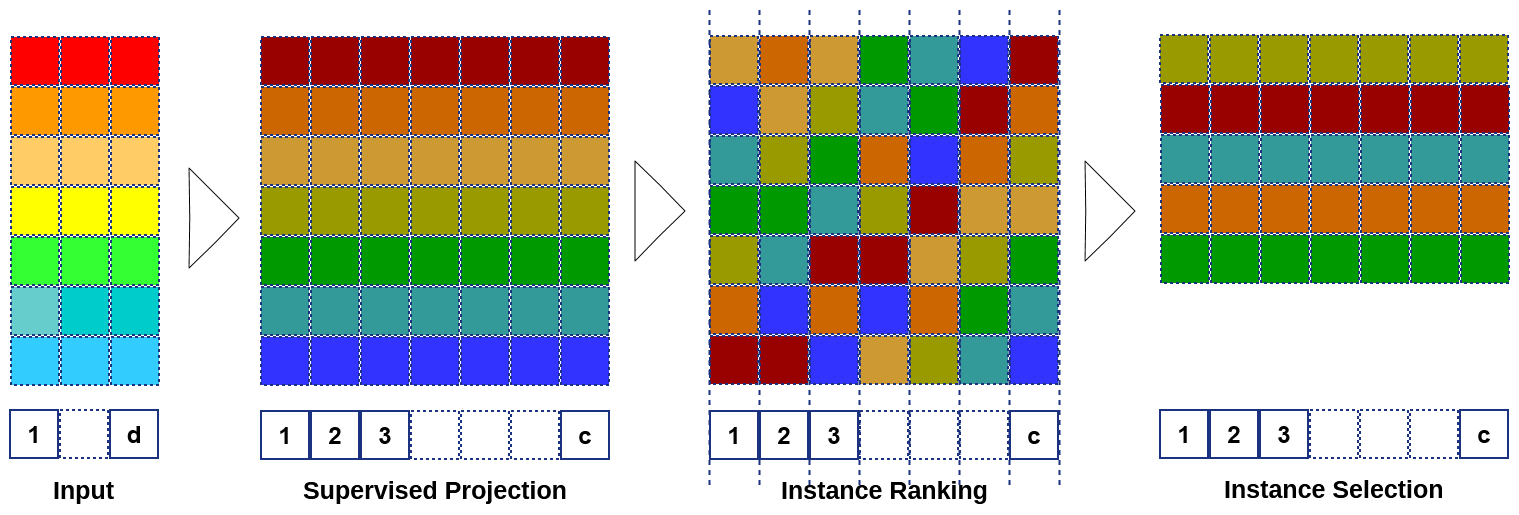}}
\caption{Our instance ranking-selection framework}
\label{fig:ranking}
\end{center}
\end{figure}

We make use of this separation as a criterion measure to rank the projected instances. Because each individual dimension exposes a specific class, our method deploys multipartite ranking to score the separability of the instances. Aggregating the scores coming from each class, measures overall distinction of each of the instances. With these scores at hand, our algorithm selects high quality instances by applying adaptive thresholding. Figure~\ref{fig:ranking} visualizes our multipartite instance ranking-selection framework in more details.

We conduct our experiments on texture understanding task. Texture is an important visual clue for various tasks in scene understanding such as material recognition~\cite{hu2011toward}\cite{timofte2012training}, texture perception and description~\cite{cimpoi2015deep}, segmentation and synthesis~\cite{liu2012sorted}. On the other hand, texture features are computed by convolution of raw images with a bank of filters. They are highly correlated and their length (dimension) corresponds to the number of color channels which are three for color images and one for grayscale samples. For an average texture dataset, the number of pixels becomes thousands of millions that means, its feature set would be an array with millions of three/one-dimensional instances. 

A huge number of above instances are redundant due to similarity between different classes of textures. Hence, it seems that employing a well-crafted instance ranking-selection algorithm will improve performance of any texture classifiers. It is worth mentioning that our framework is general and can be applied to any set of features. 

To conclude the introduction, we point out three main contributions of our work. The first contribution is supervised projection from low to high dimensions in contrast with common dimensional reduction practices in the literature. The second contribution is using bipartite ranking to perform multipartite scoring due to the fact that each of the projected instances contains the information of all available classes which are recorded in its higher dimensions. The third contribution is employing adaptive thresholding for instance selection to avoid processing of irrelevant or redundant instances. 

We organize the paper as follows: Section~\ref{framework} introduces our learning framework, Section~\ref{projection} and Appendix~\ref{appendix1} formulate our supervised projection and Section~\ref{instance} elaborates our instance ranking-selection framework. We proceed by reporting our experiments in Section~\ref{experiments} and finally, concluding in Section~\ref{conclusion}.

\section{Multipartite Ranking-Selection Framework}
\label{framework}

Assume a set of low-dimensional instances $\mathbf{R}\in\mathbb{R}^{N\times d}$ from a dataset with $c$ labels such that $d<c$ and the instances are highly correlated. We are able to classify $\mathbf{R}$ directly but the number of false matches would probably be high and degrade the recognition precision dramatically. To reflect the notion of labels into the instances for ranking purpose, we project $\mathbf{R}$ to a space spanned by the number of classes such that separability between them gets maximized whilst scattering within them becomes minimized. 

We will elaborate this projection in more detail later but for now, we present it as a matrix $\mathbf{A}\in\mathbb{R}^{d\times c}$ and multiply it by $\mathbf{R}$ to form a projected instance set $\mathbf{V}$ as follows

\begin{eqnarray}
& \mathbf{V}^{N\times c} = \mathbf{R}^{N\times d}\times\mathbf{A}^{d\times c} & \nonumber\\
& N\gg{c}\,{,}\;\; d<c &
\label{eq:01}
\end{eqnarray}

Our goal is to learn an optimal projection $\mathbf{A^{\ast}}$ to impose the highest possible distinction in $\mathbf{V}$ such that applying a ranking-selection algorithm improves the performance of multi-label classification on the dataset under study.

\subsection{Supervised Projection}
\label{projection}

Suppose that $\mathbf{R}$ contains $N$ instances in $c$ different classes. Considering Fisher criterion~\cite{bishop2006pattern}, we aim at minimizing the ratio of inter/intra class scatterings $\mathbf{S_w}$ and $\mathbf{S_b}$ by figuring out $\mathbf{A}$ such that

\begin{equation}
{arg\,min}\;\mathcal{H(\mathbf{A})} = \frac {tr({\mathbf{A}\mathbf{S_w}\mathbf{A^T}})}{tr({\mathbf{A}\mathbf{S_b}\mathbf{A^T}})} + \lVert\mathbf{I}-{\mathbf{A}\mathbf{A^T}}\rVert_2
\label{eq:02}
\end{equation}

Here, $tr(.)$ is the trace operator, $\mathbf{I}$ indicates the identity matrix and $\lVert.\rVert_2$ corresponds to the L2-norm. The first term of Equation~\ref{eq:02} aims at making the highest possible separability among instance classes. The second part is a regularization term imposing orthogonality into the projection matrix.

As projection matrix $\mathbf{A}$ in Equation~\ref{eq:01} belongs to $\mathbb{R}^{d\times{c}}$, the scattering set $\{\mathbf{S_w},\mathbf{S_b}\}$ should be included in $\mathbb{R}^{c\times c}$ to make Equation~\ref{eq:02} dimensionally consistent. This means that we are not able to employ classic discriminant analysis~\cite{fukunaga1990statistical} to solve Equation~\ref{eq:01} because $c>d$ and the problem can not be solved by a dimensional reduction method due to the fact that the dimension of instances will be increased by projecting them to higher dimensions. 

Our solution is introducing new scatterings to expose the number of classes $(c)$ as their dimensions in contrast with conventional scatterings which are defined based on the dimension of input instances $(d)$. In Appendix~\ref{appendix1}, we formulate the new definitions through Equations~\ref{eq:17}-\ref{eq:25} and prove that the set of eigenvectors corresponding to the largest $c$ eigenvalues of $\mathbf{S_w^{-1}}\mathbf{S_b}$ is a solution for this high-dimensional projection.

To solve Equation~\ref{eq:02}, we are able to start with random initialization of initial projection matrix $\mathbf{A^{(0)}}$ to come up with the optimal projection matrix $\mathbf{A^{\ast}}$. Instead, we take the above set of eigenvectors as the initial projection matrix. Although, $\mathbf{A^{(0)}}$ can be considered as an sub-optimal projection matrix, we employ it as a starting point to optimize Equation~\ref{eq:02}. It is due to the fact that Equation~\ref{eq:15} of Appendix~\ref{appendix1} is the trace-of-quotient which can be solved by generalized eigenvalue method but, Equation~\ref{eq:02} is the quotient-of-trace that requires different solution~\cite{cunningham2015linear}. 

With initial point at hand, we try to come up with a closed from gradient of Equation~\ref{eq:02} for optimization purpose. Suppose that $\mathcal{H(\mathbf{A})}$ is composed of $\mathcal{H}_{1}(\mathbf{A})$ and $\mathcal{H}_{2}(\mathbf{A})$ as follows

\begin{equation}
\mathcal{H}_{1}(\mathbf{A}) = \frac {tr({\mathbf{A}\mathbf{S_w}\mathbf{A^T}})}{tr({\mathbf{A}\mathbf{S_b}\mathbf{A^T}})}
\label{eq:03}
\end{equation}

\begin{equation}
\mathcal{H}_{2}(\mathbf{A}) = \lVert\mathbf{I}-{\mathbf{A}\mathbf{A^T}}\rVert_2
\label{eq:04}
\end{equation}

According to matrix calculus~\cite{petersen2008matrix},

\begin{equation}
\frac{\partial{tr({\mathbf{A}\mathbf{S_w}\mathbf{A^T}})}}{\partial{\mathbf{A}}} = \mathbf{A^T}(\mathbf{S_w^T}+\mathbf{S_w})
\label{eq:05}
\end{equation}

\begin{equation}
\frac{\partial{tr({\mathbf{A}\mathbf{S_b}\mathbf{A^T}})}}{\partial{\mathbf{A}}} = \mathbf{A^T}(\mathbf{S_b^T}+\mathbf{S_b})
\label{eq:06}
\end{equation}

\noindent and hence, 

\begin{eqnarray}
\frac{\partial{\mathcal{H}_{1}}}{\partial{\mathbf{A}}} & = & \frac{\mathbf{A^T}(\mathbf{S_b^T}+\mathbf{S_b})\times{tr({\mathbf{A}\mathbf{S_w}\mathbf{A^T}})}}{\big({tr({\mathbf{A}\mathbf{S_b}\mathbf{A^T}})}\big)^2} \nonumber\\
&-& \frac{\mathbf{A^T}(\mathbf{S_w^T}+\mathbf{S_w})\times{tr({\mathbf{A}\mathbf{S_b}\mathbf{A^T}})}}{\big({tr({\mathbf{A}\mathbf{S_b}\mathbf{A^T}})}\big)^2}
\label{eq:07}
\end{eqnarray}

On the other hand,

\begin{equation}
\frac{\partial{\mathcal{H}_{2}}}{\partial{\mathbf{A}}} = \frac{\partial({\mathbf{I}-{\mathbf{A}\mathbf{A^T}})}}{\partial{\mathbf{A}}}\times\frac{\mathbf{I}-{\mathbf{A}\mathbf{A^T}}}{\lVert\mathbf{I}-{\mathbf{A}\mathbf{A^T}}\rVert_2}
\label{eq:08}
\end{equation}

\noindent which gives 

\begin{equation}
\frac{\partial{\mathcal{H}_{2}}}{\partial{\mathbf{A}}} = \frac{-2\mathbf{A^T}\times(\mathbf{I}-{\mathbf{A}\mathbf{A^T}})}{\lVert\mathbf{I}-{\mathbf{A}\mathbf{A^T}}\rVert_2}
\label{eq:09}
\end{equation}

\begin{algorithm}[t]
   \caption{Supervised Projection}
   \label{alg:projection}
\begin{algorithmic}
   \STATE {\bfseries Input:} instance set $\mathbf{R}\in\mathbb{R}^{N\times d}$
   \STATE {\bfseries Output:} optimal projection matrix $\mathbf{A^{\ast}}\in\mathbb{R}^{d\times c}$ and projected instance set $\mathbf{V^{\ast}}$
   \STATE
   \STATE 1. Compute $\mathbf{S_w}$~(Eq.\ref{eq:17}) and $\mathbf{S_b}$~(Eq.\ref{eq:25})
   \STATE 2. Set $\mathbf{A^{(0)}} = eigen(\mathbf{S_w^{-1}}\mathbf{S_b})$ 
   \STATE 3. Optimize Equation~\ref{eq:02} to compute $\mathbf{A^{\ast}}$
   \STATE 4. Deploy Equation~\ref{eq:01} to obtain projected instance set $\mathbf{V^{\ast}}$
\end{algorithmic}
\end{algorithm}

Among a variety of solvers for optimization problem of Equation~\ref{eq:02}, we employ the Fast Iterative Shrinkage-Thresholding Algorithm (FISTA)~\cite{beck2009fast} which is a gradient descent method with mathematical proof of fast convergence. For implementation, we utilize UnLocBox toolbox~\cite{combettes2011proximal}. Algorithm~\ref{alg:projection} summarizes the procedure of computing optimal projection $\mathbf{A^{\ast}}$ and projected instance set $\mathbf{V^{\ast}}$.

\subsection{Instance Ranking-Selection}
\label{instance}

After projecting $\mathbf{R}$ by $\mathbf{A^{\ast}}$ to instance set $\mathbf{V^{\ast}}$, we need to employ a ranking-selection strategy. This is due to the fact that we operate on pixel level and number of instances is quite huge to be handled in a reasonable amount of time. Our target is choosing a minimal subset of instances by some criteria which removes irrelevant and redundant ones and hence, the dataset would be a better representative of data distribution. 

There are number of techniques in literature that deal with feature selection such as generating randomized subset of features directed by a classifier, sequential feature selection, using ensemble methods (bagged decision trees) and finally, ranking features by class separability criteria~\cite{liu2012feature}.

We focus on the ranking by class separability criteria because, we already introduce a class-spanned projection in Section~\ref{projection} based on Fisher criterion. Here, the challenge is how to tailor this method for instance ranking-selection. 

Considering the projected instance set $\mathbf{V^{\ast}}\in\mathbb{R}^{N\times c}$ that $N$ is the number of instances and $c$ stands for the number of classes in the dataset, there is a conceptual difference between instance and feature selection. Feature selection aims at pruning redundancy from columns of $\mathbf{V^{\ast}}$ but instance selection removes irrelevant rows. In other word, we try to tailor a column-based feature ranking algorithm to our row-based instance selection problem. 

This ranking scheme basically employs an absolute value two-sample t-test with pooled variance estimate as an independent evaluation criterion for the sake of binary classification. For multi-class sets, it deploys one versus all ranking which means holding one class and merging the others to simulate a binary labeling regime for the algorithm. This finds the proper feature columns in train set and then, select correspondent ones at test set to form the new instances.

It is impossible to adopt this strategy for instance selection because of two reasons. First, there is no correspondence between instance rows in the train and test sets. Second, we do not know labels at the test time and hence, we are not able to apply independent feature ranking for the test set.

Our solution to address this problem is based on the orthogonality imposed by the second term of Equation~\ref{eq:02} in our supervised projection paradigm. This orthogonality lets us suppose that each column of projected instance set $\mathbf{V^{\ast}}$ corresponds directly to each individual class in the dataset. This holds for both train and test sets, because we learn the optimal projection $\mathbf{A^{\ast}}$ from former and apply it to the latter. It means that the notion of labels is exposed to the test set although we do not know them at the test time.

Hence, we consider spanned classes in columns of $\mathbf{V^{\ast}}$ as pseudo-labels and start from the first column which corresponds to the first pseudo-label. We merge the remaining $c-1$ columns as a single class and employ the above feature ranking algorithm to give the ranks and criterion values for instances in the first column of $\mathbf{V^{\ast}}$. We repeat the same procedure for the second column (pseudo-label) and so forth until we come up with $c$ criterion measures for each $N$ projected instances (columns) of $\mathbf{V^{\ast}}$. 

For overall ranking, we score above projected instances by summing the measures for each individual instance. The highest the score, the better separated instance in the projected set $\mathbf{V^{\ast}}$. With overall scores at hand, we are able to select high-ranked instances and prune the rest. 

\begin{algorithm}[t]
   \caption{Instance Ranking-Selection}
   \label{alg:instance}
\begin{algorithmic}
   \STATE {\bfseries Input:} projected instance set $\mathbf{V^{\ast}}\in\mathbb{R}^{N\times c}$
   \STATE {\bfseries Output:} set of selected instances
   \STATE
   \FOR{$i=1$ {\bfseries to} $c$}
   \STATE Set $group(i) = 1$ and $group(\sim i) = 0$
   \STATE Set $criterion(1:N,i)=rank(\mathbf{V^{\ast}},group)$
   \ENDFOR
   \STATE
   \STATE Set $score(1:N) = \sum_{i=1}^{c}criterion(1:N,i)$
   \STATE Set $threshold = OTSU(score)$
   \STATE
   \FOR{$j=1$ {\bfseries to} $N$}
	  \IF{$score(j) < threshold$} 
	  \STATE Remove Instance $j$ from $\mathbf{V^{\ast}}$
	  \ENDIF
  \ENDFOR
\end{algorithmic}
\end{algorithm}

There are several selection strategies that can be applied to the overall scores. We can either select the top $m<N$ instances or use a predefined threshold to prune them. Another strategy is adaptive thresholding that we deploy Otsu's method~\cite{otsu1975threshold} for this purpose because of the fact that it is roughly a one-dimensional discrete analog of Fisher discriminant analysis. Algorithm~\ref{alg:instance} represents a pseudo code for our instance ranking-selection method.

\section{Experiments}
\label{experiments}

In our experiments, we employ three well-known local texture descriptors (LM, MR and Schmid filter banks) and apply our framework to four publicly available texture datasets (KTH-TIPS2-a, KTH-TIPS2-b, FMD and DTD). We also deploy the Oxford Visual Geometry Group's implementation~\cite{cimpoi14describing}. It reports mean accuracy of texture recognition averaged over standard number of splits according to the evaluation protocols.

We consider three filter banks consisting of 99 filters with size $49\times{49}$. The first filter bank is Leung-Malik (LM)~\cite{leung2001representing} includes 36 first and second derivatives of Gaussian filters at three scales $\{\sqrt{2},2,2\sqrt{2}\}$ and six orientations $\{\frac{\pi}{6},\frac{\pi}{3}\dots,\pi\}$, eight LoG and four Gaussian filters at scales $\{\sqrt{2},2,2\sqrt{2},4\}$. The second filter bank is Maximum-Response (MR)~\cite{varma2003texture} includes 36 filters at three scales $\{1,2,4\}$ and six orientations added to two isotropic Gaussian and LoG filters. The third filter bank is Schmid (S)~\cite{schmid2001constructing} contains 13 rotationally invariant filters with $\sigma\in\{2,4,6,8,10\}$ and $\tau\in\{1,2,3,4\}$.

We also conduct our experiments on the following texture datasets. KTH-TIPS2-a and KTH-TIPS2-b~\cite{mallikarjuna2006kth} stand for Textures under varying Illumination, Pose and Scale which the latter consists of 4572 images (4 samples, 108 images per sample and 11 categories) and the former uses only 72 images for 4 out of 44 samples. We try \cite{timofte2012training} which images on one material sample are used to train and the other three samples to test. Flicker Material Dataset (FMD)~\cite{sharan2009material} includes 1000 images (100 images per category, 10 categories) selected manually from Flickr. We follow \cite{sharan2009material} on evaluation by using 50 images per class for training and remaining 50 for testing. Describable Texture Dataset (DTD)~\cite{cimpoi2014describing} contains 5640 annotated texture images with one or more adjectives in a vocabulary of 47 English words (120 representative images per attribute). There are 10 preset splits into equally-sized training, validation and test sets.

\subsection{Experiment 1}
\label{experiment1}

Assume a set of filters $\mathcal{F}=\{\mathbf{F_1},\mathbf{F_2},\dots,\mathbf{F_{|F|}}\}$ such that filter $\mathbf{\mathbf{F_\mathit{k}}}$ is the $\mathit{k}$th filter of the bank. To start, we convert texture images  $\mathcal{I}=\{\mathbf{I_1},\mathbf{I_2},\dots,\mathbf{I_{|I|}}\}$ to standard CIE-Lab color format and normalize them to zero mean unit variance to make set $\mathcal{I}_\mathit{Lab}$. Our framework is not constrained on the dimension of the input data, so we employ all CIE-Lab color components in contrast with only using luminance channel which is a common practice in literature. This gives the chance of deploying all information in texture luminance and chrominance channels for the purpose of recognition.

Now, we consider filter $\mathbf{F_\mathit{k}}$ and convolve it with all images in the set $\mathcal{I}_\mathit{Lab}$ to obtain a set of filter responses $\mathbf{R}\in\mathbb{R}^{N\times{3}}$ such that

\begin{equation}
\mathbf{R_\mathit{k}} = \mathbf{\mathcal{I}_\mathit{Lab}}\ast\mathbf{F_\mathit{k}}
\label{eq:10}
\end{equation}

Then, we apply our supervised learning algorithm of Section~\ref{alg:projection} to compute  $\mathbf{A^{\ast}_\mathit{k}}\in\mathbb{R}^{3\times{c}}$ where $3$ is the number of color components for each pixel and $c$ stands for the number of texture classes. This projection matrix is multiplied by $\mathbf{R_\mathit{k}}$ to form a projected instance set $\mathbf{V^{\ast}_\mathit{k}}$ as follows

\begin{equation}
\mathbf{V^{\ast}_\mathit{k}} = \mathbf{R_\mathit{k}}\times\mathbf{A^{\ast}_\mathit{k}}
\label{eq:11}
\end{equation}

Here, we employ our instance ranking-selection method to prune $\mathbf{V^{\ast}_\mathit{k}}$ such that the final ensemble of projected instances, improves the performance of texture recognition task. We repeat above procedure for each filter of set $\mathcal{F}$ in parallel because, this learning process is independent for each individual filter of the bank. After computing projected instances for all filters, we create an ensemble of features $\mathcal{V^{\ast}}=\{\mathbf{V^{\ast}_1},\mathbf{V^{\ast}_2},\dots,\mathbf{V^{\ast}_{|F|}}\}$ and follow the successful practice of dictionary learning for texture understanding~\cite{cimpoi2015deeparXive}.

In Table~\ref{tab-projection}, we show our performance on texture classification in terms of mean accuracy for our instance ranking-selection framework. The first column represents Improved Fisher Vector (IFV)~\cite{cimpoi2015deeparXive} as baseline, the second one provides performances on its combination with our selected instances (Ours) and the third column shows the percentage of improvement over baseline (\%Gain) respectively.

\begin{table}[tb]
\begin{center}
\caption{Mean accuracy of texture recognition with ranking}
\label{tab-projection}
\begin{tabular}{lccc}
\hline\noalign{\smallskip}
Dataset & IFV & IFV(Ours) & \%Gain \\
\noalign{\smallskip}
\hline
\noalign{\smallskip}
KTH-a & $82.5\pm5.3$ & $\mathbf{86.4\pm4.0}$ & $\mathbf{+\%4.7}$ \\
KTH-b & $70.8\pm2.7$ & $\mathbf{75.7\pm2.7}$ & $\mathbf{+\%6.9}$ \\
FMD   & $59.8\pm1.6$ & $\mathbf{84.9\pm1.3}$ & $\mathbf{+\%42.0}$ \\
DTD   & $58.6\pm1.2$ & $\mathbf{60.5\pm1.3}$ & $\mathbf{+\%3.2}$ \\
\hline
\end{tabular}
\end{center}
\end{table}

According to the results, our framework performs highly competitive on FMD with nearly $\%42$ improvement in the precision of texture recognition followed by $\%6.9$ for KTH-TIPS2-b, $\%4.7$ for KTH-TIPS2-a and $\%3.2$ for DTD datasets. It is worth noting that KTH datasets consist quality texture images captured on controlled lighting conditions and fix distances~\cite{mallikarjuna2006kth} hence, better improvements in comparison to DTD are expected. In spite of huge difference in the number of classes, our framework does a quite competitive job on DTD, although it is far from the performance on FMD dataset.

These improvements are due to the fact that our learning framework highly separates texture classes in the proposed class space. It is also worth mentioning that this works quite competitive on datasets of texture images with various number of instances. It is also computationally efficient because we learn a supervised projection rather than the whole texture filter itself and hence, easily expands for large number of filters that can be learned in parallel for better generalization.

\subsection{Experiment 2}
\label{experiment2}

In this experiment, we try to optimize the filter banks with respect to our supervised projection scheme. Suppose that the filter $\mathbf{\mathbf{F_\mathit{k}}}$ is generated by a real function $\mathbf{f}_\mathit{k}(.)$ which is generally Gaussian or Laplacian of Gaussian (LoG). This function deploys scale $(s_\mathit{k})$, orientation $(o_\mathit{k})$, and resolution $(r_\mathit{k})$ to provide the filter matrix $\mathbf{\mathbf{F_\mathit{k}}}$ of size $r_\mathit{k}\times r_\mathit{k}$ as follows

\begin{equation}
\mathbf{\mathbf{F_\mathit{k}}} = \mathbf{f}_\mathit{k}(s_\mathit{k},o_\mathit{k},r_\mathit{k})
\label{eq-01}
\end{equation}

Following the same practice of Section~\ref{experiment1}, we go through Equations~\ref{eq:10} and~\ref{eq:11} but here, we try to minimize the Fisher criterion for $\mathbf{V^{\ast}_\mathit{k}}$ assuming its inter/intra class scatterings as $\{\mathbf{S^{\ast}_w},\mathbf{S^{\ast}_b}\}$. These are computed by Equations~\ref{eq:17}-\ref{eq:25} to define an optimization problem as

\begin{equation}
{arg\,min}\;\mathcal{P_F(\mathbf{F_\mathit{k}})} = \frac {tr(\mathbf{S^{\ast}_w})}{tr(\mathbf{S^{\ast}_b})} 
\label{eq-27}
\end{equation}

Convolution is a linear operator but filter $\mathbf{F_\mathit{k}}$ is a nonlinear function of $\{s_\mathit{k},o_\mathit{k},r_\mathit{k}\}$ according to Equation~\ref{eq-01}. Hence, above minimization problem is a nonlinear optimization with respect to the filter parameters. Besides, this minimization problem suffers from lack of generalization which might lead to ill-conditioned scattering matrices. To tackle above challenges, we redefine Equation~\ref{eq-27} as a least-square minimization problem

\begin{eqnarray}
&{arg\,min}\;\mathcal{P^{\ast}_F({\mathbf{s}_\mathit{k},\mathbf{o}_\mathit{k},\mathbf{r}_\mathit{k}})} =& \nonumber\\ 
&\bigg(log\big[tr(\mathbf{S^{\ast}_w})\big]\bigg)^2
+\bigg(\dfrac{1}{log\big[tr(\mathbf{S^{\ast}_b})\big]}\bigg)^2
+\bigg(1-log\bigg[\dfrac{tr(\mathbf{S^{\ast}_w})}{tr(\mathbf{S^{\ast}_b})}\bigg]\bigg)^2
\label{eq-28}
\end{eqnarray}

We consider the first two terms in $\mathcal{P^{\ast}_F}$ as smoothing functions which impose such a symmetry to $\mathcal{P_F}$ that avoids biases towards majority texture classes. Besides, logarithm function improves the overall convergence rate. The solution of Equation~\ref{eq-28} is the set of optimal filter parameters $\{\mathbf{s^{\ast}_\mathit{k}},\mathbf{o^{\ast}_\mathit{k}},\mathbf{r^{\ast}_\mathit{k}}\}$ that finally provides the optimal texture filter $\mathbf{F^{\ast}_\mathit{k}}$ for the convolution. We wrap up this optimization process in Algorithm~\ref{alg:optimization}. To solve Equation~\ref{eq-28}, we employ nonlinear least-squares minimization with trust-region-reflective algorithm and use the built-in implementation of Matlab optimization toolbox~\cite{coleman1996reflective}.

\begin{algorithm}[t]
   \caption{Filter Optimization}
   \label{alg:optimization}
\begin{algorithmic}
   \STATE {\bfseries Input:} set of selected instances
   \STATE {\bfseries Output:} optimal filter parameters $\{\mathbf{s^{\ast}_\mathit{k}},\mathbf{o^{\ast}_\mathit{k}},\mathbf{r^{\ast}_\mathit{k}}\}$
   \STATE
   \FOR{$k=1$ {\bfseries to} $|F|$}
   \STATE Set $\mathbf{F_\mathit{k}^{(0)}} = \mathbf{F_\mathit{k}}$~(Equation~\ref{eq-01}) 
   \STATE Compute $\mathbf{S^{\ast}_w}$~(Equation~\ref{eq:17}), $\mathbf{S^{\ast}_b}$~(Equation~\ref{eq:25})
   \STATE Solve Equation~\ref{eq-28} by trust-region-reflective algorithm
   \ENDFOR
\end{algorithmic}
\end{algorithm}

Table~\ref{tab-optimization} presents our performance on texture recognition with optimal filters. It can be seen that our framework improves the performance over ranking-only on all the datasets under study. For KTH-TIPS2-a and KTH-TIPS2-b, the improvements related to learning of filter parameters added to the ranking-only experiment, are $\%0.5$ and $\%3.8$ respectively. On FMD and DTD datasets, we get almost the same improvement with respect to the baseline as previous experiment which means, learning of filter parameters can add $\%0.2$ and $\%0.9$ to our performance.

\begin{table}
\begin{center}
\caption{Mean accuracy of texture recognition with ranking and filter optimization}
\label{tab-optimization}
\begin{tabular}{lccc}
\hline\noalign{\smallskip}
Dataset & IFV & IFV(Ours) & \%Gain \\
\noalign{\smallskip}
\hline
\noalign{\smallskip}
KTH-a & $82.5\pm5.3$ & $\mathbf{86.8\pm3.8}$ & $\mathbf{+\%5.2}$ \\
KTH-b & $70.8\pm2.7$ & $\mathbf{78.4\pm2.5}$ & $\mathbf{+\%10.7}$ \\
FMD   & $59.8\pm1.6$ & $\mathbf{85.0\pm1.2}$ & $\mathbf{+\%42.2}$ \\
DTD   & $58.6\pm1.2$ & $\mathbf{66.7\pm1.4}$ & $\mathbf{+\%13.8}$ \\
\hline
\end{tabular}
\end{center}
\end{table}

But DTD shows $\%10.6$ improvement over previous gain after filter optimization. The number of classes in DTD dataset is almost five times of the other datasets hence, the discrimination power of our supervised projection is not solely enough to separate the details of similar texture classes and tailoring the filter parameters based on the complexity of each dataset, performs significantly better for large number of classes.

\section{Conclusion}
\label{conclusion}

In this paper, we propose a novel instance ranking-selection framework targeting low-dimensional instances and apply it for the purpose of texture understanding which is a hard challenge in pattern recognition. Our scheme consists of supervised projection to a high-dimensional space, using multipartite scoring in this space for instance ranking and employing adaptive thresholding for selection to prune irrelevant or redundant instances with no contribution to the proposed recognition task. Our experiments on several texture datasets confirm the efficiency of our framework to make significant improvements in accuracy compared to the state-of-the-art local texture descriptors.

\newpage

\appendix
\section{Projection}
\label{appendix1}

To explain our specific interpretation of projection to higher dimensions, we start to formulate a classical dimension reduction method and extend it to our proposed projection paradigm.

Given $n$ samples of dimension $d$ in set $\mathbf{X}=\{x_1,x_2,\ldots,x_{n}\}$ in $\mathbb{R}^{d\times{n}}$, we propose to find a matrix $\mathbf{\overleftarrow{A}}\in\mathbb{R}^{d\times{r}}$ that maps each input vector $x_i\in\mathbb{R}^{d\times{1}}$ onto the point ${y}_i=\mathbf{\overleftarrow{A}^T}x_i$ in a lower dimensional space $\mathbb{R}^{r\times{1}}$ conditioned on $r\ll{d}$. We try to maximize separability between and minimize scattering within classes of set $\mathbf{X}$. 

One of the most popular methods for recovering this mapping with supervised learning is linear discriminant analysis (LDA)~\cite{fukunaga1990statistical}. Here, the mapping matrix $\mathbf{\overleftarrow{A}}$ is determined to minimize the Fisher criterion given by

\begin{equation}
	\mathcal{J_F}(\mathbf{\overleftarrow{A}})=tr\bigg((\mathbf{\overleftarrow{A}^T}\mathbf{\overleftarrow{S}_w}\mathbf{\overleftarrow{A}})(\mathbf{\overleftarrow{A}^T}\mathbf{\overleftarrow{S}_b}\mathbf{\overleftarrow{A}})^{-1}\bigg)
\label{eq:15}
\end{equation}

\noindent which $\mathit{tr(.)}$ is diagonal summation operator. The within-class scattering $\mathbf{\overleftarrow{S}_w}\in\mathbb{R}^{d\times{d}}$ is defined as

\begin{equation}
	\mathbf{\overleftarrow{S}_w} = \sum_{j=1}^c\sum_{x_i \in\mathbf{C}_j}(x_i - \mu_j)(x_i - \mu_j)^T
\label{eq:16}
\end{equation}

\noindent and the between-class scattering $\mathbf{\overleftarrow{S}_b}\in\mathbb{R}^{d\times{d}}$ as

\begin{equation}
	\mathbf{\overleftarrow{S}_b} = \sum_{j=1}^c(\mu_j-\bar{\mu})(\mu_j-\bar{\mu})^T
\label{eq:17}
\end{equation}

\noindent where $c$, $\mu_j$ and $\bar{\mu}$ are number of classes, mean over class $\mathbf{C}_j$ and mean over all dataset respectively. 

The matrix $\mathbf{\overleftarrow{S}_w}$ can be regarded as the average class-specific covariance, whereas $\mathbf{\overleftarrow{S}_b}$ can be viewed as the mean distance between all different classes. Thus, the purpose of Equation \ref{eq:01} is to maximize the between-class scatter while preserving within-class dispersion in the mapped space. 

The $\mathbf{\overleftarrow{A}}$ computes by solving a generalized eigenvalue problem like $\mathbf{\overleftarrow{S}_b}\mathbf{\overleftarrow{A}}= \lambda\mathbf{\overleftarrow{S}_w}\mathbf{\overleftarrow{A}}$~\cite{bishop2006pattern}. Since rank of $\mathbf{\overleftarrow{S}_b}$ is $r-1$, the solution is eigenvectors corresponding to the largest $r-1$ eigenvalues of $\mathbf{\overleftarrow{S}_w^{-1}}\mathbf{\overleftarrow{S}_b}$ for $r\ll{d}$.

Assuming $\mathbf{\overleftarrow{S}_b}\neq\mathbf{I}$ and $\mathbf{\overleftarrow{S}_w}\neq\mathbf{I}$, by cyclic permutation of trace operator, the Equation~\ref{eq:15} holds

\begin{eqnarray}
    \mathcal{J_F}(\mathbf{\overleftarrow{A}}) &=& tr\bigg((\mathbf{\overleftarrow{A}^T}\mathbf{\overleftarrow{S}_w}\mathbf{\overleftarrow{A}})(\mathbf{\overleftarrow{A}^T}\mathbf{\overleftarrow{S}_b}\mathbf{\overleftarrow{A}})^{-1}\bigg)\nonumber\\
    &=& tr\bigg(\mathbf{\overleftarrow{A}^T}\mathbf{\overleftarrow{S}_w}\mathbf{\overleftarrow{A}}(\mathbf{\overleftarrow{A}})^{-1}\mathbf{\overleftarrow{S}_b^{-1}}(\mathbf{\overleftarrow{A}^T})^{-1}\bigg)\nonumber\\
	&=& tr\bigg((\mathbf{\overleftarrow{A}}^T)^{-1}\mathbf{\overleftarrow{A}^T}\times\mathbf{\overleftarrow{S}_w}\times\mathbf{I}\times\mathbf{\overleftarrow{S}_b^{-1}}\bigg)\nonumber\\
	&=& tr(\mathbf{\overleftarrow{S}_w}\mathbf{\overleftarrow{S}_b^{-1}})
\label{eq:18}
\end{eqnarray}

\noindent that here, $\mathbf{I}$ is the identity matrix. For a non-invertible matrix, Moore-Penrose pseudo-inverse~\cite{penrose1955generalized} is a common generalized inverse based on SVD factorization but here, there is no need to compute any inverses.

To come up with our proposed projection for $c$ classes of samples ($c>d$) in the set $\mathbf{X}$, we again consider the Fisher criterion in Equation \ref{eq:15} and define new inter scattering $\mathbf{\overrightarrow{S}_w}\in\mathbb{R}^{c\times{c}}$ such that satisfies

\begin{equation}
	tr(\mathbf{\overleftarrow{S}_w}) = tr(\mathbf{\overrightarrow{S}_w})
\label{eq:19}
\end{equation}

Note that in Equations~\ref{eq:16}, we sum over all classes $(c)$ and hence, to satisfy Equation~\ref{eq:19}, we can consider $\mathbf{\overrightarrow{S}_w}$ as a square matrix of size $c\times{c}$ with all entries equal zero except main diagonals as  

\begin{equation}
	\mathbf{\overrightarrow{S}_w}(j,j) = tr\bigg(\sum_{x_i\in \mathbf{C}_j}(x_i-\mu_j)(x_i-\mu_j)^T\bigg) \quad\forall j\in [1,c]
\label{eq:20}
\end{equation}

From Equations~\ref{eq:19} and similarity invariance of trace operator, $\mathbf{\overleftarrow{S}_w}\in\mathbb{R}^{d\times{d}}$ and $\mathbf{\overrightarrow{S}_w}\in\mathbb{R}^{c\times{c}}$ are similar matrices~\cite{horn2012matrix} which implies, there should exist a non-singular matrix $\mathbf{\Gamma_w}\in\mathbb{R}^{c\times{d}}$ such that

\begin{equation}
	\mathbf{\overleftarrow{S}_w} = \mathbf{\Gamma_w^{-1}}\mathbf{\overrightarrow{S}_w}\mathbf{\Gamma_w}
\label{eq:21}
\end{equation}

By minor matrix operations, Equation~\ref{eq:21} arranges as

\begin{equation}
	\mathbf{\Gamma_w}\mathbf{\overleftarrow{S}_w} - \mathbf{\overrightarrow{S}_w}\mathbf{\Gamma_w} = \mathbf{0}
\label{eq:22}
\end{equation}

\noindent which is a special case of Sylvester equation~\cite{lee2011simultaneous} for square matrices $\{\mathbf{\overleftarrow{S}_w},\mathbf{\overrightarrow{S}_w}\}$ and can be solved for $\mathbf{\Gamma_w}$ by either Kronecker tensor trick or using generalized eigen decomposition because, we define $\mathbf{\overleftarrow{S}_w}$ and $\mathbf{\overrightarrow{S}_w}$ as non-singular matrices. The closed form solution for Equation~\ref{eq:22} by Roth's removal rule~\cite{gerrish1998sylvester} is

\begin{equation}
	vec(\mathbf{\Gamma_w}) = \mathbf{I}\otimes({- \mathbf{\overrightarrow{S}_w}})-(\mathbf{\overleftarrow{S}_w})^T\otimes\mathbf{I}
\label{eq:23}
\end{equation}

\noindent which $vec(.)$ is vectorization operator and $\otimes$ is Kronecker product. With the same reasoning, we define $\mathbf{\overrightarrow{S}_b}$ as a square matrix of size $c\times{c}$ such that

\begin{equation}
	tr(\mathbf{\overleftarrow{S}_b}) = tr(\mathbf{\overrightarrow{S}_b})
\label{eq:24}
\end{equation}

\noindent and there should exist a non-singular matrix $\mathbf{\Gamma_b}\in\mathbb{R}^{c\times{d}}$ such that

\begin{equation}
	\mathbf{\overleftarrow{S}_b} = \mathbf{\Gamma_b^{-1}}\mathbf{\overrightarrow{S}_b}\mathbf{\Gamma_b}
\label{eq:25}
\end{equation}

On the other hand, from Equations~\ref{eq:21} and \ref{eq:25}

\begin{eqnarray}
    \mathbf{\overleftarrow{S}_w}\mathbf{\overleftarrow{S}_b^{-1}} &=& (\mathbf{\Gamma_w^{-1}}\mathbf{\overrightarrow{S}_w}\mathbf{\Gamma_w}) (\mathbf{\Gamma_b^{-1}}\mathbf{\overrightarrow{S}_b}\mathbf{\Gamma_b})^{-1}\nonumber\\
    &=& \mathbf{\Gamma_w^{-1}}\mathbf{\overrightarrow{S}_w}\mathbf{\Gamma_w}
    \mathbf{\Gamma_b^{-1}}\mathbf{\overrightarrow{S}_b^{-1}}\mathbf{\Gamma_b}
\label{eq:26}
\end{eqnarray}

Due to the similarity invariance in Equations~\ref{eq:19} and \ref{eq:24}, we consider the cyclic permutation of trace operator and assign

\begin{equation}
    \mathbf{\Gamma_b}=\mathbf{\Gamma_w} 
\label{eq:27}
\end{equation}

As a result, Equation~\ref{eq:25} implies $\mathbf{\overrightarrow{S}_b}$ as

\begin{equation}
	\mathbf{\overrightarrow{S}_b} = \mathbf{\Gamma_b}\mathbf{\overleftarrow{S}_b}\mathbf{\Gamma_b^{-1}}
\label{eq:28}
\end{equation}

Now, we work out Equation~\ref{eq:26} by substitution from Equation~\ref{eq:27} as follows

\begin{eqnarray}
    \mathbf{\overleftarrow{S}_w}\mathbf{\overleftarrow{S}_b^{-1}}
    &=& \mathbf{\Gamma_w^{-1}}\mathbf{\overrightarrow{S}_w}\mathbf{\Gamma_w}
        \mathbf{\Gamma_w^{-1}}\mathbf{\overrightarrow{S}_b^{-1}}\mathbf{\Gamma_w}\nonumber\\
    &=& \mathbf{\Gamma_w^{-1}}\mathbf{\overrightarrow{S}_w}\times \mathbf{I}\times\mathbf{\overrightarrow{S}_b^{-1}}\mathbf{\Gamma_w}\nonumber\\
	&=& \mathbf{\Gamma_w^{-1}}(\mathbf{\overrightarrow{S}_w}\mathbf{\overrightarrow{S}_b^{-1}})\mathbf{\Gamma_w}
\label{eq:29}
\end{eqnarray}

This proves $\mathbf{\overleftarrow{S}_w^{-1}}\mathbf{\overleftarrow{S}_b}$ and $\mathbf{\overrightarrow{S}_w^{-1}}\mathbf{\overrightarrow{S}_b}$ are similar matrices such that it holds

\begin{equation}
    tr(\mathbf{\overleftarrow{S}_w}\mathbf{\overleftarrow{S}_b^{-1}}) = tr(\mathbf{\overrightarrow{S}_w}\mathbf{\overrightarrow{S}_b^{-1}})
\label{eq:30}
\end{equation}

Looking back at Equation~\ref{eq:18}, we are now able to define a new optimization problem for $\mathbf{\overrightarrow{S}_w}$ and $\mathbf{\overrightarrow{S}_b}$ considering the same discrimination power and projection orthogonality of Equation~\ref{eq:15} as

\begin{eqnarray}
    \mathcal{J_F}(\mathbf{\overrightarrow{A}}) &=& tr\bigg((\mathbf{\overrightarrow{A}}\mathbf{\overrightarrow{S}_w}\mathbf{\overrightarrow{A}^T})(\mathbf{\overrightarrow{A}}\mathbf{\overrightarrow{S}_b}\mathbf{\overrightarrow{A}^T})^{-1}\bigg)
\label{eq:31}
\end{eqnarray}

\noindent which is aligned with the number of classes $(c)$ instead of dimension of input $(d)$. Employing the same eigenvector solution as Equation~\ref{eq:15} to minimize Equation~\ref{eq:31}, gives the projection matrix $\mathbf{\overrightarrow{A}}\in\mathbb{R}^{d\times{c}}$ for $c>d$.


\newpage

\bibliography{arxiv2016}
\bibliographystyle{splncs03}

\end{document}